\documentclass[preprint,12pt]{elsarticle}

\biboptions{square,numbers}
\usepackage{lineno,hyperref}
\modulolinenumbers[1]

\journal{~}

\usepackage[utf8]{inputenc} 
\usepackage[T1]{fontenc}    
\usepackage{hyperref}       
\usepackage{url}            
\usepackage{booktabs}       
\usepackage{amsfonts}       
\usepackage{nicefrac}       
\usepackage{microtype}      
\usepackage{xcolor}

\usepackage{soul}
\usepackage{graphicx}
\def\<{\ensuremath{\langle}}
\def\>{\ensuremath{\rangle}}
\def\x{\ensuremath{{\bf x}}}
\def\y{\ensuremath{{\bf y}}}
\def\z{\ensuremath{{\bf z}}}
\def\s{\ensuremath{{{\bf s}}}}

\def\M{\ensuremath{{\cal M}}}

\def\flip{\ensuremath{{\textsf{flip}}}}

\def\pos{\ensuremath{{\textsf{pos}}}}

\def\max{\ensuremath{{\textsf{max}}}}
\def\min{\ensuremath{{\textsf{min}}}}
\usepackage{xspace}
\usepackage{xcolor}
\usepackage{colortbl}
\usepackage{amsmath,amsfonts,amsthm}
\usepackage{latexsym} 
\usepackage{amssymb} 
\newtheorem{example}{Example}
\newtheorem{definition}{Definition}
\newtheorem{theorem}{Theorem}

\newtheorem{problem}{Problem}
\def\miocolore{black!10}
\def\mycolor{white}
\usepackage{setspace}
\usepackage[noend]{algorithmic}
\def\titolo{Even-if Explanations: \\ Formal Foundations, Priorities and Complexity}

\usepackage{multirow}
\usepackage{enumitem}
\title{\titolo}
\usepackage{setspace} 
\usepackage{algorithm}

\usepackage{setspace}

\begin{document}

\begin{frontmatter}

\title{Even-if Explanations: \\ Formal Foundations, Priorities and Complexity}

\author{Gianvincenzo Alfano, Sergio Greco, Domenico Mandaglio, \\ Francesco Parisi,  Reza Shahbazian, and  Irina Trubitsyna}

\address{Department of Informatics, Modeling, Electronics and System Engineering,\\
University of Calabria, Italy\\[4pt]
\{g.alfano, greco, d.mandaglio, fparisi, i.trubitsyna\}@dimes.unical.it reza.shahbazian@unical.it}

\begin{abstract}
Explainable AI has received significant attention in recent years. 
Machine learning models often operate as black boxes, 
lacking explainability and transparency 
while supporting decision-making processes. 
Local post-hoc explainability queries attempt to answer why individual inputs are classified in a certain way by a given model. 
While there has been important work on counterfactual explanations,
less attention has been devoted to semifactual ones. 
In this paper, we focus on local post-hoc explainability queries 
within the semifactual `even-if' thinking 
and their computational complexity among different classes of models,
and show that both linear and tree-based models are 
strictly more interpretable than neural networks.
After this, we introduce a preference-based framework 
enabling users to personalize explanations based on their preferences, 
both in the case of semifactuals and counterfactuals, 
enhancing interpretability and user-centricity. 
Finally, we explore the complexity of several interpretability 
problems in the proposed preference-based framework 
and provide algorithms for polynomial cases.
\end{abstract}

\end{frontmatter}

\section{Introduction}
The extensive study of counterfactual `if only' thinking, exploring how things might have been different, has been a focal point for social and cognitive psychologists
\cite{kahneman1981simulation,mccloy2002semifactual}.
Consider a negative event, such as taking a 
taxi
and due to traffic arriving late to a party. By analyzing this situation, an individual (e.g.  Alice) might engage in counterfactual thinking by imagining how things could have unfolded differently, such as, `if only Alice had not taken the taxi, she would not have arrived late at the party'.
This type of counterfactual thinking, where an alternative scenario is imagined, is a common aspect of daily life. 
In such a case the counterfactual scenario negates both the event's cause (antecedent) and its outcome, presenting a false cause and a false outcome that are temporarily considered as true (e.g., Alice took the taxi and arrived late).

Counterfactual thinking forms the basis for crafting counterfactual explanations,  which are crucial in automated decision-making processes. 
These explanations leverage imagined alternative scenarios, aiding users in understanding why certain outcomes occurred and how different situations might have influenced decisions. Counterfactual explanations empowers users to grasp the rationale behind decisions, fostering transparency and user trust in these systems. 
Several definitions of counterfactual explanations exist in 
the literature~\cite{guidotti2022counterfactual}. 
According to most of the literature, counterfactuals are defined as the minimum changes to apply to a given instance to let the prediction of the model be different~\cite{NIPS20}. 

\begin{figure}[t]
\centering
\includegraphics[scale=.42]{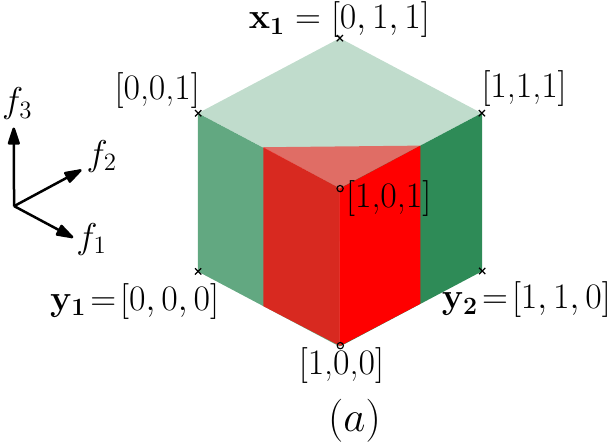}\hspace*{3mm}
\includegraphics[scale=.72]{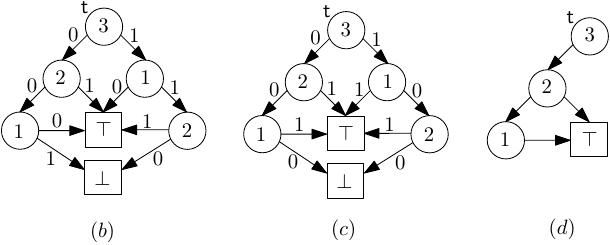}
\caption{(a) Binary classification model $\M: step(\x\cdot [-2,2,0]+1)$ of Example~\ref{ex:intro1} representing the hiring scenario.  
The binary feature $f_1$ (resp.,  $f_2$ and $f_3$) represents part-time employment contract (resp.,  salary lower than 5K\$, and on site-working).   
Crosses (resp.,  circles) on the corners of the green (resp.,  red)  area correspond to instances where the model outputs $1$ (resp., $0$).
(b) FBDD model $\M=(V,E,\lambda_V,\lambda_E)$ of Example~\ref{ex:FBDD} with root $\textsf{t}$ and $\lambda_V(\textsf{t})=3$.  (c) FBDD model $\M'=(V',E',\lambda_{V'},\lambda_{E'})$ computed at Line 1 of Algorithm~\ref{alg:FBDD}.
(d) Graph $\cal N$ obtained at Line 2 of Algorithm~\ref{alg:FBDD}. 
Squared nodes represent leaf nodes ($\top$ for $\M(\cdot)=1$, and $\bot$ for $\M(\cdot)=0$).
}
\label{fig:intro1} 
\end{figure}

While significant attention in AI has been given to counterfactual explanations, there has been a limited focus on the equally important and related semifactual `even if' {explanations}~\cite{ijcai2023p732,kennyHuangEvenIf},  though they have been investigated much more in cognitive sciences.
While counterfactuals explain what changes to the input features of an AI system change the output decision,  semifactuals show which input feature changes 
do not change a decision outcome.
Considering the above-mentioned situation 
where Alice took the taxi and arrived late at the party,
we might analyze `even if' scenarios envisioning how things could have remained the same, such as, ``even if Alice had not taken a taxi, she would have still arrived late at the party''.

Sharing the same underlying idea of counterfactuals, we define semifactuals as the maximum changes to be
applied to a given instance while keeping the same prediction. 
Indeed, the larger the feature differences asserted in the semifactual, the better (more convincing) the explanation~\cite{ijcai2023p732}. This intuitively captures the desire of an agent to have more degree of freedom and favorable conditions (represented by features changed), while keeping the (positive) status assigned to it by the model.
As an example,  consider the following,  inspired by an hiring scenario. 

\begin{example}\label{ex:intro1}\rm
Consider the binary and linear classification model $\M: \{0,1\}^3\rightarrow \{0,1\}$ shown in Figure~\ref{fig:intro1}(a), where $\M$ is defined as $step(\x\cdot [-2,2,0]+1)$ and the input $\x = [x_1,x_2,x_3]$ denotes an applicant (also called user) defined by means of the following three features:
\begin{itemize}
\item
$f_1=$``part-time job''
\item 
$f_2=$``requested (monthly) salary < 5K\$''
\item
$f_3=$``on-site job''.
\end{itemize}
For any instance $\x\in \{0,1\}^3$ we have that $\M(\x)=0$ if $\x=[1,0,1]$ or $\x=[1,0,0]$,  and $\M(\x)=1$ otherwise.  
Intuitively,  this means that the company's AI model does not approve the application only when the user applies for a part-time job and the requested salary is greater than or equal to 5K\$.

Consider a user $\x_1$ that applies for a full-time and on-site job,  and  the requested salary is lower than 5K\$ (i.e., $\x_1=[0,1,1]$), we have that $\y_1=[0,0,0]$ and $\y_2=[1,1,0]$ are semifactual of $\x_1$ w.r.t.  $\M$ at maximum distance (i.e., 2) from $\x_1$ in terms of number of feature changed.
Intuitively, $\y_1$ represents the fact that `the user $\x_1$ will  be hired \textit{even if}   (s)he would have requested for a remote job and the requested salary would have been greater than or equal to 5K\$', while $\y_2$ represents `the user $\x_1$ will  be hired \textit{even if}  (s)he would have applied for a remote and part-time job'.~\hfill~$\Box$
\end{example}

However, as highlighted in the previous example, multiple semifactuals can exist for each given instance. 
In these situations,  a user may prefer one semifactual to another, by expressing preferences over features so that the \textit{best} semifactuals will be selected, as shown in the following example.

\begin{example}\label{ex:intro2}\rm
Continuing with Example~\ref{ex:intro1}, suppose that the user 
$\x_1$ looks for another opportunity and  prefers to change feature  $f_2$ rather than $f_1$ 
(irrespective of any other change), that is (s)he prefers semifactuals with $f_2=0$ rather than those with $f_1=1$.
Thus, (s)he would prefer to still get hired by changing the salary to be greater than or equal to 5K\$  (obtaining $\y_1$); if this cannot be accomplished, then (s)he prefers to 
get it by changing the job to part-time (obtaining $\y_2$).~\hfill~$\Box$
\end{example}

Prioritized reasoning in AI,  focusing on incorporating user preferences, represents a pivotal advancement in the field, enhancing  
adaptability and user-centricity of AI systems.  
Traditional AI models rely on predefined rules or optimization criteria to generate outcomes, often overlooking the nuanced nature of user-specific preferences \cite{2011Rossi,2016Santhanam}.
Prioritized reasoning addresses this limitation by introducing a mechanism that allows users to express their preferences, thereby guiding AI systems to prefer  specific factors over others in the decision-making processes.

One key aspect of prioritized reasoning is its applicability across diverse AI domains, spanning machine learning~\cite{kapoor2012performance}, natural language processing~\cite{bakker2022fine}, and recommendation systems~\cite{zhu2022personalized}.
In machine learning, for instance, the ability to prioritize specific features or outcomes based on user preferences significantly improves the relevance and usability of the resulting models. 
Within natural language processing, prioritized reasoning facilitates customized language generation aligned with individual preferences, catering to diverse communication styles.
The impact of prioritized reasoning extends to recommendation systems, where user preferences play a crucial role in shaping the suggestions provided. 

Our work contributes to prioritized reasoning within explainable AI in the presence of user's preference conditions related to features.  
These preferences are exploited to generate semifactual and counterfactual explanations that align most closely with the user-specified criteria.
In particular, preferences are applied similarly to what has been proposed  in the  well-known 
Answer Set Optimization (ASO) approach for Logic Programs with preferences~\cite{BrewkaNT03}.

\newpage
\paragraph{Contributions}
Our main contributions are as follows.
 \begin{itemize}[leftmargin=5mm,itemsep=-2pt]
\item 
We   formally introduce the concepts of semifactual over three classes of models: 
$(i)$ \textit{perceptrons} $(ii)$ \textit{free binary decision diagrams (FBBDs)},  and $(iii)$ \textit{multi-layer perceptrons (MLP)}, intuitively encoding local post-hoc explainable queries within the even-if thinking setting.
Herein, the term `local' refers to explaining the output of the system 
for a particular input, while `post-hoc'
refers to interpreting the system after it has been trained.
 \item 
We investigate the computational complexity of 
interpretability problems concerning semifactuals,  
showing that 
they are not more difficult than analogous
problems related to counterfactuals~\cite{NIPS20}.
 \item 
We introduce a framework that empowers users to prioritize explanations according to their subjective preferences. That is, users can specify preferences for altering specific features over others within explanations. This approach enriches the explanation process, enabling users to influence the selection of the most favorable semifactuals (called ``best'' semifactuals), 
thereby augmenting the interpretability and user-centricity of the resulting outputs. 
Notably, the proposed framework also naturally encompasses preferences over counterfactuals. 
\item 
We investigate the 
complexity of several interpretability
problems related to best semifactuals and best counterfactuals.  
Table~1 summarizes our complexity results.
Finally, focusing on a restricted yet expressive class 
of feature preferences, we identify tractable cases for which 
we propose algorithms for their computation.
\end{itemize}

\section{Preliminaries}\label{sec:prel}

We start by recalling the key concepts underlying counterfactual and semifactual explanations,  and then we recall 
the  main complexity classes used in the paper.

\paragraph{Classification Models} A (binary classification) model is a function  ${\cal M} :  \{0, 1\}^n \rightarrow \{0, 1\}$, specifically focusing on {instances whose features are represented by binary values.} Constraining inputs and outputs to booleans simplifies our context while encompassing numerous relevant practical scenarios. A class of models is just a way of grouping models together.
An instance $\bf x$  is a vector in $\{0, 1\}^n$ and represents a possible input for a model. 
We  now recall three significant categories of ML models that will be the ones we will focus on.

\noindent  
A \textit{Binary Decision Diagram} (BDD) $\M=(V,E,\lambda_V,\lambda_E)$~\cite{WEGENER2004229} is a rooted directed acyclic graph $(V,E)$ where $(i)$ leave nodes are either labeled 1 (also denoted as $\top$) or 0 (also denoted as $\bot$), $(ii)$  internal nodes are labeled by function $\lambda_V$ with a value from $\lbrace 1, \dots, n \rbrace$,  and $(iii)$ each internal node has two outgoing edges labeled by function $\lambda_E$ as $1$ and $0$,  respectively.
Each instance ${\bf x} = [x_1,\dots,x_n] \in  \{0,1\}^n$ uniquely maps to a path $p_{\bf x}$ in $\cal M$. 
This path adheres to the following condition: for every non-leaf node $u$ in $p_{\bf x}$ labeled $i$, the path goes through the edge labeled with $x_i$. 
$|\mathcal{M}|$ denotes the size of $\mathcal{M}$, representing the number of edges. A binary decision diagram $\cal M$ is \textit{free} (FBDD) if for every path from the root to a leaf, no two nodes on that path have the same label. 
A \textit{decision tree} is simply 
an FBDD whose underlying graph is a tree.

\noindent A \textit{multilayer perceptron (MLP)} $\cal M$ with $k$ layers is defined by a sequence of weight matrices ${\bf W}^{(1)},\dots , {\bf W}^{(k)}$,  bias vectors ${\bf b}^{(1)},\dots, {\bf b}^{(k)}$,  and activation functions {$a^{(1)}, \dots, a^{(k)}$}. 
Given an instance $\bf x$, we inductively define 
{${\bf h}^{(i)} = a^{(i)} ({\bf h}^{(i-1)} {\bf W}^{(i)} + {\bf b}^{(i)})$ with  $i\in \{1,\dots, k\}$},  
assuming that ${\bf h}^{(0)} = {\bf x}$.  
The output of $\cal M$ on $\bf x$ is defined as 
${\cal M}({\bf x}) = {\bf h}^{(k)}$. 
In this paper we assume all weights and biases to be rational numbers, i.e. belonging to $ \mathbb{Q}$. 
We say that an MLP as defined above has $(k- 1)$ \textit{hidden layers}. 
The \textit{size} of an MLP $\cal M$, denoted by $|\cal M|$,  is the total size of its weights and biases, in which the size of a rational number $\frac{p}{q}$ is $log_2(p)+log_2(q)$ (with the convention that $log_2(0) = 1$).
We focus on MLPs in which all internal functions {$a^{(1)},\dots, a^{(k-1)}$}  are the ReLU function $relu({x}) = max( 0,  x)$. 
Usually,  MLP binary classifiers are trained using the sigmoid as the output function {$a^{(k)}$}.
Nevertheless,  when an MLP classifies an input (after training), it takes decisions by simply using the preactivations,  also called logits. 
Based on this and on the fact that we only consider already trained MLPs,  we can assume w.l.o.g.  that the output function {$a^{(k)}$} is the binary \textit{step} function,  defined as $step({x}) = 0$ if $x<0$,  
and $step({x}) = 1$  if $x\geq 0$.

\noindent A \textit{perceptron} is an MLP with no hidden layers (i.e.,  $k = 1$). 
That is,  a perceptron $\cal M$ is defined by a pair $({\bf W}, b)$ 
such that ${\bf W}\in \mathbb{Q}^{n\times 1}$  and $b\in \mathbb{Q}$, and the output is ${\cal M}({\bf x}) = step({\bf x}\, {\bf W}+b)$. 
Because of its particular structure,  a perceptron is usually defined as a pair $({\bf w},  b)$ with $\bf w=W^T$ a rational vector and $b$ a rational number.  
The output of ${\cal M}(\bf x)$ is then $1$ 
iff  ${\bf x \cdot w} + b \geq 0$,  where $\bf x \cdot w$ denotes the dot product between $\bf x$ and $\bf w$.

\paragraph {Complexity Classes}
Boolean functions $\cal{F}$ mapping strings to strings whose output is a single bit are called decision problems.  We identify the computational problem of computing $\cal{F}$  (i.e., given an input string $x$ compute ${\cal{F}}(x)$)  with the problem of deciding whether ${\cal{F}}(x)=1$.
As an example of $\cal{F}$, consider to take as input a graph $G$ and a number $k$ and return $1$ whether there exists a  set of vertices {each-other not adijacent} of size at least $k$.   
This decision problem is {known as  \textsc{Independent Set}.
We recall the complexity classes used in the  paper and,  in particular, the definition of the classes PTIME (or, briefly, P),  NP and coNP (see e.g. ~\cite{book-Papadimitriou}).  
P (resp., NP)  contains the set of decision problems that can be solved in polynomial time by a deterministic (resp.,  nondeterministic) Turing machine.  Moreover,  coNP is the complexity class containing the complements of problems in NP.
A problem $\mathcal{P}$ is called hard for a complexity class $C$ (denoted $C$-hard) if there exists a {polynomial} reduction from any problem in $C$ to $\mathcal{P}$. 
If $\mathcal{P}$ is both $C$-hard and belongs to $C$ then it is $C$-complete.
Clearly,  P $\subseteq$ NP,  P $\subseteq$ coNP and, under standard theoretical assumptions,  the three classes are assumed to be different.

\section{Even-if Explanations}\label{sec:semifactual}
In this section, we instantiate our framework on three important classes of boolean models and explainability queries. Subsequently, we present our main theorems, facilitating a comparison of these models in terms of their interpretability.
We start by recalling the notion of counterfactual, that is the explainability notion in the `if only' case,  whose complexity has been investigated in~\cite{NIPS20}.  
We define the \textit{distance measure} between two instances $\x, \y \in \lbrace 0,1 \rbrace^n$ as the number of features where they differ.  Formally,  $d(\x,\y)=\sum_{i=1}^{n}|x_i-y_i|$ is the number of indexes 
$i\in\{1,\dots,n\}$ (i.e., features) where $\x$ and $\y$ differ.

\begin{definition}[Counterfactual]
Given a pre-trained model $\M$ and an instance $\x$,  an instance $\y$ is said to be a counterfactual of $\x$ iff $i)$ $\M(\x)\neq \M(\y)$,  and
$ii)$ there exists no other instance $\z\!\neq\!\y$ s.t.  $\M(\x)\neq \M(\z)$ and $d(\x,\z)\!<\!d(\x,\y)$. 
\end{definition}

\begin{example}\label{ex:counter1}\rm
Continuing with our running example illustrated in Figure~\ref{ex:intro1}(a),  
for $\y_3=[1,0,1]$ we have that $\x_2=[0,0,1]$ and $\x_3=[1,1,1]$ are  the only counterfactuals of $\y_3$ w.r.t. $\M$ (herein, $d(\y_3,\x_2)=d(\y_3,\x_3)=1$).
Intuitively, this encodes the fact that user $\y_3$ (that applied for a part-time and remote job, and a salary greater than or equal to 5K\$) will be hired \textit{if only} (s)he would change the employment contract to be full time (obtaining $\x_2$) or the requested salary to be lower than 5K\$ (obtaining $\x_3$).~\hfill~$\Box$
\end{example}

The natural decision version of the problem of finding a couterfactual for $\x$ is the following.

\begin{problem} [\cite{NIPS20}] $[$\textsc{Minimum Change Required (MCR)}$]$ Given a model $\mathcal{M}$, instance $\bf x$, and $k\in \mathbb{N}$,  check  whether there exists an instance $\bf y$ with  $d(\x,\y)\leq k$ and $\mathcal{M}(\bf x)\neq \mathcal{M}(\bf y)$.
\end{problem}

\begin{theorem}[\cite{NIPS20}]\label{thm:MCR}
\textsc{MCR}  is $i)$ in PTIME for FBDDs and perceptrons,  and  $ii)$ 
NP-complete for MLPs. 
\end{theorem}

We follow a standard assumption about the relationship between model  interpretability and computational complexity~\cite{NIPS20}: 
\textit{a class ${\cal A}$ of models is more interpretable than another class ${\cal B}$ if the computational complexity of addressing post-hoc queries for models in ${\cal B}$ is higher than for those in ${\cal A}$}. 
Under this assumption, Theorem \ref{thm:MCR} states  that the class of models `perceptron' and `FBDD' is strictly more interpretable than the class `MLP', as the computational complexity of answering post-hoc queries for models in the first two classes is lower than for those in the latter.
These results represent a principled way to confirm the folklore belief that linear models are more interpretable than deep neural networks within the context of interpretability queries for counterfactuals.

An open question is whether the same holds when dealing with post-hoc queries based on the 'even-if' thinking setting, i.e. on semifactuals. 
Before exploring this research question, we formally introduce the concept of semifactual.

\begin{definition}[Semifactual]\label{def:semi}
Given a pre-trained model $\M$ and an instance $\x$, an instance $\y$ is said to be a semifactual of $\x$ iff $i)$ $\M(\x)= \M(\y)$, and $ii)$ there exists no other instance $\z\!\neq\!\y$ s.t. $\M(\x) = \M(\z)$ and  $d(\x,\z)\!>\!d(\x,\y)$. 
\end{definition}

Similar to counterfactuals, the following problem  is the  decision version of the problem of finding a semifactual of an instance $\x$  with a model $\M$.

\begin{problem}$[$\textsc{Maximum Change Allowed (MCA)}$]$ Given a model $\mathcal{M}$, instance $\bf x$, and $k\in \mathbb{N}$,  check  whether there exists an instance $\bf y$ with $d(\x,\y)\geq k$ and $\mathcal{M}(\bf x)= \mathcal{M}(\bf y)$.
\end{problem}

	Although semifactuals and counterfactuals appear to be similar, 
	their mathematical definitions are different.  
	Indeed, while counterfactuals minimize the changes in order to have a different outcome, 
	semifactuals maximize the changes while keeping the same outcome.  
	Notably, the two problems are not interchangeable - 
	we do not see how to naturally reduce
	one to the other; however, a (possibly complex) reduction
	may exist as our complexity results presented in Theorem~\ref{thm:MCA} below do not rule this out.  
	For instance, considering Example~\ref{ex:intro1} and the two semifactuals $\y_1$ and $\y_2$ of $\x_1$, they do not correspond to the counterfactuals of the counterfactuals of $\x_1$, that are $[0,0,1]$ and $[1,1,1]$.

\begin{theorem}\label{thm:MCA}
\textsc{MCA}  is $i)$ in PTIME for FBDDs and perceptrons,  and $ii)$ NP-complete for MLPs. 
\end{theorem}

\noindent 
{\em Proof sketch.}
(Perceptron).~\textsc{MCA} (optimization) problem under perceptrons can be formulated in ILP as the standard 
(max-) Knapsack problem where each item has value $1$ 
and weight $w_i (2x_i - 1)$,
and the bag has capacity $\sum_{i=1}^n w_i x_i$.  
The Knapsack problem is, in the 
general case, NP-Hard~\cite{book-Papadimitriou}.
However,  \textsc{MCA} corresponds to a special instance of Knapsack where every item has the same cost that can be solved in polynomial time with a greedy strategy, from which the result follows. \\
\noindent
(FBDD).~Let $\M_u$ be the FBDD obtained by restricting $\M$ to the nodes that are (forward-)reachable from $u$ and $mca_u(\x)=\max\{k' \mid \exists \y.\ d(\x,\y) = k'\wedge \M_u(\y)\text{=}\M(\x)\}$ 
(that can be comptuted in PTIME).
For $\y$ maximizing $k'$ it holds that $y_{u'} \neq x_{u'}$ holds $\forall u'$ from the root of $\M$ to $u$ excluded.
Let $r$ be the root of $\M$.  
Then,  we can show that $(\M,\x,k)$ is a positive  instance of \textsc{MCA} iff $mca_r(\x) \geq k$. \\ 
\noindent
(MLP)
The following algorithm provides the membership in NP.  
Guess an instance $\y$ and check~in~PTIME that $d(\x,\y)$ $\geq k$ and $\M(\x)=\M(\y)$.
We prove the hardness with a polynomial reduction from 
\textsc{Independent Set},
which is known to be NP-complete~\cite{book-Papadimitriou}.~\hfill~$\Box$

It turns out that, under standard complexity assumptions,  
computing semifactuals under perceptrons and FBDDs is easier than under multi-layer perceptrons.  
Moreover, independently of the type of the model, 
computing semifactuals is as hard as  computing counterfactuals
(cf. Table~1). 
Thus,  perceptrons and FBDDs are strictly {more} interpretable than MLPs,  in the sense that the complexity of answering post-hoc queries for models in the first two classes is lower than for those in the latter. 
 
\begin{table}[t!]\label{tab:complexity}
\footnotesize
\begin{center}
\begin{tabular}{p{1cm}|c|c|c|c|c|c|}  
\cline{2-7} 
   & \textsc{MCR} &  \textsc{MCA}   & \textsc{cb-MCR} &  \textsc{cb-MCA} & \textsc{cbl-MCR} &  \textsc{cbl-MCA} \\
\hline \hline 
\multicolumn{1}{|l||}{\!\!\! \!FBDDs\!\!\!}   & \cellcolor{\miocolore} P & \cellcolor{\mycolor} P  & \cellcolor{\mycolor}  coNP & \cellcolor{\mycolor}  coNP & P & P  \\ \hline   
\multicolumn{1}{|l||}{\!\!\! \!Perceptrons\!\!\!}   & \cellcolor{\miocolore} P & \cellcolor{\mycolor} P  & \cellcolor{\mycolor}  coNP & \cellcolor{\mycolor}  coNP & P & P  \\ \hline    
\multicolumn{1}{|l||}{\!\!\! \!MLPs\!\!\!}   & \cellcolor{\miocolore} NP-c &   \cellcolor{\mycolor} NP-c  & \cellcolor{\mycolor}  coNP-c & \cellcolor{\mycolor}  coNP-c & \cellcolor{\mycolor}  coNP-c & \cellcolor{\mycolor}  coNP-c \\ \hline     
\end{tabular}
\end{center}
\caption{Complexity of explainability queries for models of the form $\{0,1\}^n\rightarrow \{0,1\}$.
For any class C, C-c  
 means C-complete. 
Grey-colored cells refer to known results.} 
\end{table}

\section{Preferences over Explanations}\label{sec:prefer}
The problem of preference handling has been extensively studied in AI. Several formalisms have been proposed to express and reason with different kinds of preferences~\cite{BrafmanD09,2011Rossi,2016Santhanam}. 
In the following, in order to express preferences over semifactuals and counterfactuals, we introduce a novel approach inspired to that proposed in \cite{BrewkaNT03}, whose semantics is based on the \textit{degree} to which preference rules are satisfied. 
 
\noindent{\bf Syntax.} Input instances are of the form $\x = (x_1,...,x_n)$ in $\{0,1\}^n$.
Each $x_i$ (with $i \in [1,n]$) represents the value of feature $f_i$, and the (equality) atom $f_i = x_i$ denotes that the value of feature $f_i$ in $\x$ is equal to $x_i$.
A \emph{simple preference} is  an expression of the form $(f_i = x_i') \succ (f_j = x_j'')$; it intuitively states that we prefer instances (i.e. semifactuals or counterfactuals) where  $f_i = x_i'$ w.r.t. those where $f_j = x_j''$.
To simplify the notation, 
an equality atom of the form $f_i = 1$ is written as a positive atom $f_i$, whereas an atom of the form $f_i = 0$ is written as a negated atom $\neg f_i$. Positive and negated atoms are also called (feature) literals.

\begin{definition}[Preference rule]\rm
Let ${\cal I} = \{ f_1,...,f_n \}$ be the set of input features.
A preference rule,  where $m\geq k \geq 2$, and any $\varphi_{i}\in \{f_1,\neg f_1,\dots f_n,\neg f_n\}$ is a (feature) literal, with $i\in [1,m]$ has the following form:
 
$ \ \ \ \ \ \ \varphi_{1} \succ \cdots \succ \varphi_{k} \leftarrow \varphi_{{k+1}} \wedge \cdots \wedge \varphi_{m} \ \ \ \ (1)$. 
\end{definition}

In (1), $\varphi_{1} \succ \cdots \succ \varphi_{k}$ is called head (or consequent), whereas $\varphi_{k+1} \wedge \cdots \wedge \varphi_{m}$ is called body (or antecedent).
We assume that literals in the head are distinct.
Intuitively, whenever $\varphi_{k+1}, \cdots, \varphi_{m}$ are true, then $\varphi_{1}$ is preferred over $\varphi_{2}$,   which is preferred over $\varphi_{3}$,  
and so on until $\varphi_{k-1}$ which is preferred over $\varphi_{k}$.
As usual, when the body of the rule is empty, the implication symbol $\leftarrow$ is omitted.

\begin{definition}[\textsc{BCMp} framework]\rm
A (binary classification) model with preferences (\textsc{BCMp}) framework is a pair $(\M,\succ)$ where $\M$ is a model and $\succ$\footnote{With a little abuse of notation, 
we use $\succ$ to denote both a set of preference and the preference relation among feature literals.} a set of preference rules over 
features of $\cal{M}$. 
\end{definition}

A practical and natural way for users to express their preferences in our framework includes---but are not limited to---specifying a ranking on a (sub)set of features whose values users would prefer to change (or keep unchanged).

\begin{example}\label{ex:syntax}\rm
The \textsc{BCMp} framework $\Lambda_1 = (\M, \ \{\neg f_2\!\succ\! f_1\})$, where $\M$ is the model presented in Example~\ref{ex:intro1}, encodes the user's preference specified in 
Example~\ref{ex:intro2}.
Consider now the framework $\Lambda_2$ obtained from $\Lambda_1$ by replacing the preference rule with $\neg f_2 \!\succ\! f_1 \leftarrow \neg f_3$. 
Roughly speaking, it encodes a user preference stating that among the explanations (i.e. counterfactuals/semifactuals) satisfying the condition that \emph{the work is remote} (i.e., $\neg f_3$ holds), the ones where \emph{the salary is greater than or equal to 5K\$} (i.e.,  $\neg f_2$ holds) are preferred and, if this is not possible, those where \emph{the work is part-time} (i.e.,  $f_1$) are taken. ~\hfill~$\Box$
\end{example}

\noindent {\bf Semantics.}
 To formally establish an ordering among explanations, a partial order $\sqsupseteq$ derived from the set of preference rules, is introduced next. 
Let us consider an explanation $\y$ and a preference $\kappa$ of the form (1),  the three situations are possible:}\\
\hspace*{-3mm} (a)  the body of $\kappa$ is not satisfied in {$\y$};\footnote{A positive (resp., negative) atom $f_i$ (resp., $\neg f_i$) is satisfied in $\y\text{=}(y_1,\dots,y_n)$ whenever $y_i\text{=}1$ (resp., $y_i\text{=}0$).}\\
\hspace*{-3mm} (b) the body of $\kappa$ is satisfied in $\y$ and at least one head literal is true in $\y$;\\
\hspace*{-3mm} (c) 
 the body of $\kappa$ is satisfied in $\y$ and none of the head literals is satisfied in $\y$.

In  the  cases (a)  and (b) we   say  that    $\y$  \textit{satisfies}  $\kappa$  respectively  with  degree  $1$   and  $min(\{l\mid {\y}\ \text{satisfies}\ \varphi_{l}\ \text{with}\ l\leq k \})$ (denoted as $\delta({\y},\kappa)$),  while in case (c) we say that {$\y$} \textit{does not satisfy} $\kappa$ and the associated degree is $\delta({\y},\kappa)=+\infty$.
Intuitively, $\delta({\y},\kappa)$ represents the position of the first feature literal satisfied in the ordered list provided in the head of a preference rule; however, it can be $1$ if {$\y$} satisfies 
the first literal $\varphi_{1}$ or if the rule is irrelevant (case a).
If the rule is relevant (the body is satisfied) and no head literal is satisfied, then it is  $+ \infty$.

Thus,  given a \textsc{BCMp} framework ($\M,\succ$), an instance $\x$ and two explanations $\y$ and $\z$ for $\M$ and $\x$, we write $\y\sqsupseteq \z$ iff $\delta({\y},\kappa)\leq \delta({\z},\kappa)$ for all preference $\kappa$ in $\succ$, and write $\y\sqsupset \z$ iff $\y\sqsupseteq \z$ and  $\z\not \sqsupseteq \y$.

\begin{definition}[Semantics]\rm
Given a \textsc{BCMp} framework $({\cal M}, \succ\!)$ and an instance $\bf x$.  We say that $\y$ is a \textit{best semifactual} (resp., \textit{counterfactual}) explanation  of $\x$ if $\y$ is a semifactual (resp., counterfactual)  of $\x$ and there is no other semifactual (resp., counterfactual)  $\z$ of $\x$ such that  $\z\sqsupset \y$.
\end{definition}

\begin{example}\label{ex:semantics1}\rm
Continuing with Example~\ref{ex:syntax}, the instances  $\y_1=[0,0,0]$ and $\y_2=[1,1,0]$ satisfy $\kappa=\neg f_2 \!\succ\! f_1 \leftarrow \neg f_3$ with degree $\delta(\y_1,\kappa)=1$ 
and $\delta(\y_2,\kappa)=2$ since the second and first literal in the head of $\kappa$ is satisfied in $\y_1$ and $\y_2$, respectively, and the body of $\kappa$ is satisfied by both of them. 
Then,  $\y_1\sqsupset \y_2$, and thus $\y_1$ is the only best semifactual of $\x_1=[0,1,1]$.
~\hfill~$\Box$
\end{example}

\noindent{\bf Computational Complexity. }
We now investigate the  complexity of several problems related to prioritized reasoning for explanations in order to compare model classes, even under prioritized reasoning.
Observe that deciding the \textit{existence of a best explanation} in the even-if and if-only thinking follows from deciding the existence of counterfactuals and semifactuals, respectively. Thus, preferences do not make the existence problem harder.  
 Consider now the problem of \emph{verification of a best explanation} in both settings.

\begin{problem}$[$\textsc{check {best}-MCR (cb-MCR)}$]$ Given a \textsc{BCMp} $(\M, \succ)$,  instances $\x$,  $\y$ with $d(\x,\y)=k$, and $\M(\x) \neq \M(\y)$,  check whether there is no $\z$ with $\M(\x)\neq \M(\z)$ and either $d(\x,\z)\leq k-1$,   or $d(\x,\z)= k$ and $\z \sqsupset \y$.
\end{problem}

\begin{problem}$[$\textsc{check {best}-MCR (cb-MCA)}$]$ Given a \textsc{BCMp} $(\M, \succ)$,  instances $\x$,  $\y$ with $d(\x,\y)=k$, and $\M(\x) = \M(\y)$,  check whether there is no $\z$ with $\M(\x)= \M(\z)$ and either $d(\x,\z)\geq k+1$ or $d(\x,\z)= k$ and $\z \sqsupset \y$.
\end{problem}

Observe that in our semantics  for any framework $(\M,\succ)$ and instances $\y,\z$ of $\M$, deciding whether $\y\sqsupset\z$ can be done in PTIME. We use this result to prove the following theorem.

\begin{theorem}\label{thm:CBMCR}
\textsc{cb-MCR} and \textsc{cb-MCA}  are $i)$ in coNP for FBDDs and perceptrons,  and $ii)$ coNP-complete for MLPs. 
\end{theorem}

\renewcommand{\algorithmiccomment}[1]{\hfill//~#1}
 \begin{algorithm}[t] 
    \caption{ Computing a (best) semifactual for perceptrons}
    \label{alg:MCA}
    \textbf{Input}: Perceptron $\M=({\bf W},b)$,  
    instance $\x\in \{0,1\}^n$, and   linear preference $\kappa = f_{p_1} \succ \cdots \succ f_{p_l}$.\\
    \textbf{Output}: A best semifactual $\y$ for $\x$ w.r.t.  $\M$ and $\kappa$.
    
\begin{algorithmic}[1] 
\STATE Let ${\bf s}=[f_1/s_1, \dots,f_n/s_n]$ where $ \forall i\in[1,n]$,    $s_i\!=\! 2x_i w_i-w_i\ \text{if}\ \M(\x)\!=\!1,\ w_i-2x_i w_i\ \text{otherwise}$; 

\STATE Let ${\bf s}' = [f_{q_1}/s_{q_1},...,f_{q_n}/s_{q_n}]$ be the sorted version of $\bf s$ in ascending order of $s_i$;
\STATE $k\	 \text{=}\ \max(\{i \in [0,n] \mid \M(\flip(\x,\pos(\s',i))=\M(\x)\})$; $\!\!\!$
\STATE {\textbf{if}\ $k=0$\, \textbf{return}\ $\x$;}
\STATE {\textbf{if}\ $k=n$\, \textbf{return}\ $[1-x_1,\dots,1-x_n]$;}
\STATE $\y=\flip({\x},\pos(\s',k))$;
\STATE $\delta=\min(\{i\in [1,l]\mid y_{p_i}=1\}\cup\{l+1\})$;
\FOR{$i \in [1,...,\delta-1]$}
	\STATE{\textbf{if}\ $y_{p_i}=1$\ \textbf{return}\ $\y$;} 
	\STATE {Let $j =q_k\ \text{if}\ x_{p_i}=y_{p_i}$,
	$j=q_{k+1}\ \text{otherwise}$};
	\STATE $\z=\flip(\y,\{p_i,{j}\})$;
	\STATE{\textbf{if}\ $\M(\x)=\M(\z)$\ \textbf{return}\ $\z$;}
\ENDFOR
\RETURN{$\y$;}
\end{algorithmic} 
\end{algorithm}

We now focus on a simpler yet expressive class of \textsc{BCMp}s  where a fixed linear order is defined over a subset of features.

\begin{definition}[\textsc{BCMlp} framework]\rm
A (binary classification) 
model with linear preferences (\textsc{BCMlp}) framework is a pair $(\M,\succ)$ where $\M$ is a model, and $\succ$ consists of a linear preference rule, that is a single preference rule of the form (1) with empty body.
\end{definition}

\begin{example}\label{ex:linear1}\rm
Considering Example \ref{ex:syntax}, the \textsc{BCMp} framework $\Lambda_1$ is linear, whereas $\Lambda_2$ is not. ~\hfill~$\Box$
\end{example}

We use \textsc{cbl-MCA} and \textsc{cbl-MCR} to denote  \textsc{cb-MCA} and \textsc{cb-MCR} where the input \textsc{BCMp} is linear. 
We will show that these problems are in PTIME in the case of perceptrons and FBDDs. The algorithms presented below take as input a model $\M$, an instance $\x$, 
and a linear preference $\kappa$, 
and returns a best semifactual explanation in PTIME. 
When a preference is not specified, a generic semifactual at maximum distance is returned.
For the sake of the presentation, w.l.o.g.,  we assume that head literals are positive atoms (i.e., $\varphi_i\!=\!f_i$ for all $i$).

\begin{algorithm}[t]
    \caption{Computing a (best) semifactual for FBDDs}
    \label{alg:FBDD}
    \textbf{Input}: FBDD $\M\text{=}(V,E,\lambda_V,\lambda_E)$ with root \textsf{t},  
    instance $\x\!\in\! \{0,1\}^n$,  and linear preference $\kappa = f_{p_1} \succ \cdots \succ f_{p_l}$.\\
    \textbf{Output}: {A~best~semifactual~$\y$~for~$\x$~w.r.t.~$\M$~and $\kappa$.}
      
\begin{algorithmic}[1] 

\STATE Let ${\cal{M}'}\text{=}(V'\text{=}V,E'\text{=}E,\lambda_{V'}\text{=}\lambda_V,\lambda_{E'})$  where\\ $\lambda_{E'}(u,v)\text{=}1\ \text{if}\ (x_{\lambda_V(u)}\text{=}\lambda_E(u,v)),$ $\ 0\ \text{otherwise}$; 
\STATE Let $\cal{N}=\textsf{subgraph}(\M',\M(\x))$; 
\STATE Let $\Pi$ be the set of paths in $\cal N$ from $\textsf{t}$ to leaf nodes;
\FOR{$f_{p_i}\in [f_{p_1},\dots,f_{p_l}]$}
\STATE{\!\textbf{if} 
$\exists \pi\in \Pi\ \text{with}\  \y=\textsf{build}(\x,\pi)\ \text{and}\ y_{p_i}=1$ \ \ \textbf{return}	 $\y$;}
\ENDFOR
\STATE{Let $\pi$ be a path of $\Pi$ taken non-deterministically;}\STATE{\textbf{return} 
$\y=\textsf{build}(\x,\pi)$;}
\end{algorithmic}
\end{algorithm}

Algorithm~\ref{alg:MCA} is created for the perceptron model. 
Initially, a list $\s$ of pairs feature/weight is built, 
where each weight takes into account the contribution of the 
associated feature to the result (Line 1).
At Line 2 the list is sorted in ascending order of weights, giving a new list $\s'$.
Next the feature values of $\x$ are changed to get a semifactual instance $\y$.
To change a maximum number $k$ of feature values of $\x$, guaranteeing that the output of the model $\M(\x)$ does not change, the order in $\s'$ is followed.
To this end, the following functions are introduced: 
\textit{i}) $\pos(\s',i)$, computing the set of the positions in $\s$ of the first $i$ features in $\s'$, and 
\textit{ii}) $\flip(\y,B)$, with $\y = [y_1,...,y_n]$, updating every element $y_i$ such that $i \in B$ with the complementary value $1 - y_i$. Notice that $k=0$ means that $\x$ is the only semifactual for $\x$, (returned at Line 4), whereas $k=n$ means that $[1-x_1,\dots,1-x_n]$ is the only semifactual for $\x$ (returned at Line 5).
At Line 7 the degree of satisfaction of $\kappa$ by $\y$ is computed; if no feature in $\kappa$ is satisfied by $\y$ the degree is $l+1$ (standing for $+\infty$).
The next steps (Lines 8-13) search for a better semifactual instance 
(if any).
This is carried out by considering the first $\delta-1$ features in $\kappa$.
Thus, for each feature $f_{p_i}$ in $\kappa$ ($i \in [1,\delta-1]$) we have that:
\textit{i}) if $\y$ satisfies $f_{p_i}$, $\y$ is a best semifactual 
and it is returned at Line 9;
\textit{ii}) otherwise an alternative instance $\z$ satisfying feature $f_{p_i}$ is generated and, if it is a semifactual then it is returned at Line 12. In particular,  $\z$ is a semifactual instance only if $d(\z,\x) = d(\y,\x)$ and $\M(\z) = \M(\x)$. To guarantee that $d(\z,\x) = d(\y,\x)$ we either restore feature $f_{q_k}$ in $\y$ by setting it back 
to $y_{{q_k}}=x_{{q_k}}$ whenever $x_{p_i}=y_{p_i}$ (recall that the first $k$ features of $\x$ are flipped at Line 6),
or change the feature $f_{k+1}$ (i.e., $y_{k+1}\neq x_{k+1}$) whenever $x_{p_i}\neq y_{p_i}$. 
Roughly speaking, the so-obtained feature $f_j$ minimally contributes 
to determine the value $\M(\y)$ and keeps the distance equal to $k$. Finally,  if there exist no other semifactual $\z$ for $\x$ s.t. $\z \sqsupset \y$, then $\y$ is returned at Line 13.

\begin{example}\rm
Consider the model $\M=step(\x\cdot [-2,2,0]+1)$ of 
Example~\ref{ex:intro1}, the instance $\x=[0,1,1]$ and the linear preference $f_1 \succ f_2$.  
We have that $\s=[f_1/2, f_2/2, f_3/0]$, $\s'=[f_3/0, f_2/2, f_1/2]$ and $k=2$. 
Thus $\y=[0,0,0]$ with $\delta=3$.   
As $p_1=1$ and $y_{1}=x_{1}=0$,  Algorithm~\ref{alg:MCA} returns $\z= \flip(\y,[1,j])= [1,1,0]$ with $j=q_k=2$ and $\delta=1$.~\hfill~$\Box$
\end{example}

Algorithm~\ref{alg:FBDD} works with FBDD.
It starts by creating a copy $\M'$ of $\M$ where edge labels, assigned by function $\lambda_{E'}$, represent boolean values  corresponding to whether or not the edge is in line with $\x$. 
Formally, $\lambda_{E'}(u,v)=1$ if $x_{\lambda_V(u)}=\lambda_E(u,v)$; 
$0$ otherwise. 
Intuitively, let $\pi$ be a path in $\M$ from root $\textsf{t}$ to leaf nodes representing $\M(\x)$ (i.e., to $\top$ if $\M(\x)=1$, 
$\bot$ otherwise), and let $w$ be the weight of $\pi$ computed as 
the sum of (boolean) weights on the edges of $\pi$. 
By following the path $\pi$, it is possible to build a semifactual for $\x$ of distance $n-w$.
To this end, at Line 2 a graph $\cal{N}$  is built with function $\textsf{subgraph}$ 
by keeping all the paths in $\M'$ ending in $\M(\x)$ and having minimum weight. All such paths are stored in $\Pi$ at Line 3. 
Then, the algorithm checks if it is possible to build a semifactual $\y$ for $\x$ satisfying $f_{p_1}$, otherwise $f_{p_2}$, and so on.
Particularly, assuming to be at step $f_{p_i}$ for some $f_{p_i}$ in $\kappa$, if there exists a path $\pi\in \Pi$ and the feature $f_{p_i}$ can be set to $1$ in $\y$ (that is the condition of Line 5) then a best semifactual $\y$ of $\x$ is obtained from $\x$ by flipping every features $i$ of $\x$ not appearing in $\pi$ (i.e., $\lambda_V(u)\neq i$ for any node $u$ in $\pi$) or 
differing in the assignment given by $\pi$ 
(i.e., there is no edge $(u,v)\in \pi$ s.t. $\lambda_V(u)=i$ and $\lambda_{E'}(u,v)=0$).
More formally, the function $\textsf{build}(\x,\pi)$ returns the instance $\y=\textsf{flip}(\x,\{i\in[1,n]\mid \nexists (u,v)\in \pi\ \text{such that}\ \lambda_V(u)\text{=}i\ \text{and}\ \lambda_{E'}(u,v)\text{=}1\})$.
Finally, if the algorithm does not return a semifactual at Line 6,
then at Line 8 it returns a semifactual $\y$ of $\x$ that satisfies none of the $f_{p_i}$s, obtained from $\x$ through function $\textsf{build}(\x,\pi)$ where $\pi$ is a path taken non-deterministically from $\Pi$.

\begin{example}\label{ex:FBDD}\rm
Consider the FBDD $\M=(V,E,\lambda_V,$ $\lambda_E)$ in Figure~\ref{fig:intro1}(b) for the hiring scenario of Example~\ref{ex:intro1}.  
Let $\x=[0,1,1]$ and  $\kappa=f_2\succ f_1$.
For each edge $(u,v)\in E$, Figure~\ref{fig:intro1}(c) shows the value $\lambda_{E'}(u,v)$ (computed at Line 1),  while Figure~\ref{fig:intro1}(d) shows the graph $\cal N$ obtained 
by removing all the paths $\pi$ of $\M'$ whose sum of weights is greater than $1$. 
This means that semifactuals $\y$ of $\x$ are at distance $d(\x,\y)=n-1=2$.
As there exists in $\cal{N}$ the path $\pi:(u,u'),(u',u'')$ where $\lambda_V(u)=3$, $\lambda_V(u')=2$, $\lambda_V(u'')=\top$, and $\lambda_{E'}(u',u'')=1$ then it is possible to get a semifactual $\y$ from $\x$ 
with $x_2=1$. 
In fact, Algorithm~\ref{alg:FBDD} returns 
$\y=\textsf{build}(\x,\pi)=\flip(\x,\{1,3\})=[1,1,0]$ 
at Line 6.~\hfill~$\Box$
\end{example}

\begin{theorem}\label{thm:l-CBMCA}
{\textsc{cbl-MCA} and \textsc{cbl-MCR}} are {in} PTIME 
{for}
perceptrons and FBDDs.
\end{theorem}

Providing a PTIME algorithm returning a best semifactual for MLPs is unfeasible,  as backed by our complexity analysis.
However,  
as an heuristic, we could adapt
Algorithm~\ref{alg:MCA} so that the scores $s_i$s encode feature importance, 
similarly to what is done in \cite{ramon2020comparison}---we plan to explore this direction in future work.

Observe that,  although the number of (best) semifactuals is potentially exponential w.r.t. the number of features 
(it is $\binom{n}{k}$ where $k$ represents the maximum number of features changed in $\x$ to obtain semifactuals),  
Algorithms 1 and 2 can be exploited to obtain a finite representation of all semifactuals. 
For instance, all semifactuals of $\x$ in Algorithm 2 are those obtained from $\x$ by considering all paths in $\pi\in\Pi$,
that is the set $\{\textsf{build}(\x,\pi)$ such that $\pi\in\Pi\}$.

\section{Final Discussion}~\label{sec:related}

\noindent \textbf{Related work.}
Looking for transparent and interpretable models has led to the exploration of 
several explanation paradigms in eXplainable AI (XAI). 
Factual explanations~\cite{ciravegna2020human,guidotti2018survey,bodria2023benchmarking,WangKB21,ciravegna2023logic} elucidate the inner workings of AI models by delineating why a certain prediction was made based on the input data. 
Counterfactual explanations~\cite{dervakos2023choose,romashov2022baycon,albini2020relation,wu2019counterfactual,guidotti2022counterfactual}, 
delve into hypothetical scenarios, revealing alternative input configurations that could have resulted in a different prediction. 
These explanations offer insights into the model's decision boundaries and aid in understanding its behavior under varied circumstances. 
Semifactual explanations~\cite{DandlCBB23,ijcai2023p732,kenny2021generating} bridge the gap between factual and counterfactual realms by presenting feasible alterations to the input data that do not change a decision outcome. This trichotomy of explanation types contributes significantly to the holistic comprehension and trustworthiness of AI systems, catering to various stakeholders' needs for transparency and interpretability. 
Most existing works on XAI focus on proposing novel methods for generating explanations, with few addressing the computational complexity  of related problems  \cite{NIPS20,arenas2022computing,arenas2021foundations,EibenOPS23,el2023cardinality,MarzariCCF23,ordyniak2023parameterized}. However, none of these works specifically focuses on semifactuals.  
The approach proposed in \cite{DandlCBB23} aims to derive a set of semifactuals by solving an NP-hard optimization problem, for which they introduce a model-agnostic \textit{heuristic} method. 
However,   our approach differs from that in~\cite{DandlCBB23} from several standpoints: (\textit{i}) the notion of semifactual considered (as not relying on maximal distance in~\cite{DandlCBB23}),  
(\textit{ii}) the ability to express preferences (not considered in in~\cite{DandlCBB23}),  and 
(\textit{iii}) the fact that our approach is an \textit{exact} method rather than heuristic.

~\\
\noindent {\bf Limitations.} 
Our study is based on a notion of complexity-based interpretability which depends on the specific set of queries chosen, as in~\cite{NIPS20}. 
We plan to explore additional interpretability queries, particularly focusing on counting problems like quantifying the number of semifactuals/counterfactuals for specific input instances, aiming to enhance the evaluation of interpretability across the three classes of models. Furthermore, we aim to extend our research by investigating more inclusive model formats, specifically those accommodating non-binary features, thereby expanding the breadth of our analysis. 
Using other approaches to express preferences, such as those based on Pareto Optimality~\cite{BoutilierBDHP04,RossiVW04,AllenGJMR17,GoldsmithLTW08,LukasiewiczM19}, is worth investigating for future work.  
Our algorithms focus on $(i)$ Perceptrons and FBDD, and $(ii)$ to a single linear preference, though they can be easily adapted to cope with a fixed number of (consistent) linear preferences.  Devising algorithms dealing with $(i)$ MLP and $(ii)$ multiple and different preference rules deserves further investigation.

~\\
\noindent\textbf{Conclusions.} 
Our work is also motivated by the growing interest in regulating AI~\cite{buiten2019towards}, and particularly in contexts where sensitive decisions regarding humans are demanded to AI systems. 
An important example is given by the legislation proposal in the State of New York concerning the use of such systems in recruitment processes. 
The proposal emphasizes the need to explain systems' results, focusing not so much on the decision-making process itself, but rather on the outcome produced~\cite{NYT_Article}.
As this is exactly the explainability setting considered in our work,  we believe our research could inspire future developments in regulating AI.

\clearpage

\bibliography{refs}

\begin{thebibliography}{10}
\expandafter\ifx\csname url\endcsname\relax
  \def\url#1{\texttt{#1}}\fi
\expandafter\ifx\csname urlprefix\endcsname\relax\def\urlprefix{URL }\fi
\expandafter\ifx\csname href\endcsname\relax
  \def\href#1#2{#2} \def\path#1{#1}\fi

\bibitem{kahneman1981simulation}
D.~Kahneman, A.~Tversky, The simulation heuristic, National Technical
  Information Service, 1981.

\bibitem{mccloy2002semifactual}
R.~McCloy, R.~M. Byrne, Semifactual ``even if'' thinking, Thinking \& reasoning
  8 (2002) 41--67.

\bibitem{guidotti2022counterfactual}
R.~Guidotti, Counterfactual explanations and how to find them: literature
  review and benchmarking, Data Mining and Knowledge Discovery (2022) 1--55.

\bibitem{NIPS20}
P.~Barcel{\'{o}}, M.~Monet, J.~P{\'{e}}rez, B.~Subercaseaux, Model
  interpretability through the lens of computational complexity, in:
  Proceedings of Advances in Neural Information Processing Systems, 2020.

\bibitem{ijcai2023p732}
S.~Aryal, M.~T. Keane, Even if explanations: Prior work, desiderata \&
  benchmarks for semi-factual xai, in: Proceedings of International Joint
  Conference on Artificial Intelligence (IJCAI), 2023, pp. 6526--6535.

\bibitem{kennyHuangEvenIf}
E.~M. Kenny, W.~Huang, The utility of {\textquotedblleft}even
  if{\textquotedblright} semi-factual explanation to optimise positive
  outcomes, in: Thirty-seventh Conference on Neural Information Processing
  Systems, 2023.

\bibitem{2011Rossi}
F.~Rossi, K.~B. Venable, T.~Walsh, A Short Introduction to Preferences: Between
  Artificial Intelligence and Social Choice, Synthesis Lectures on Artificial
  Intelligence and Machine Learning, Morgan {\&} Claypool Publishers, 2011.

\bibitem{2016Santhanam}
G.~R. Santhanam, S.~Basu, V.~G. Honavar, Representing and Reasoning with
  Qualitative Preferences: Tools and Applications, Synthesis Lectures on
  Artificial Intelligence and Machine Learning, Morgan {\&} Claypool
  Publishers, 2016.

\bibitem{kapoor2012performance}
A.~Kapoor, B.~Lee, D.~Tan, E.~Horvitz, Performance and preferences: Interactive
  refinement of machine learning procedures, in: Proceedings of AAAI Conference
  on Artificial Intelligence, 2012, pp. 1578--1584.

\bibitem{bakker2022fine}
M.~Bakker, M.~Chadwick, H.~Sheahan, M.~Tessler, L.~Campbell-Gillingham,
  J.~Balaguer, N.~McAleese, A.~Glaese, J.~Aslanides, M.~Botvinick, et~al.,
  Fine-tuning language models to find agreement among humans with diverse
  preferences, Proceedings of Advances in Neural Information Processing Systems
  35 (2022) 38176--38189.

\bibitem{zhu2022personalized}
Y.~Zhu, Z.~Tang, Y.~Liu, F.~Zhuang, R.~Xie, X.~Zhang, L.~Lin, Q.~He,
  Personalized transfer of user preferences for cross-domain recommendation,
  in: Proceedings of the Fifteenth ACM International Conference on Web Search
  and Data Mining, 2022, pp. 1507--1515.

\bibitem{BrewkaNT03}
G.~Brewka, I.~Niemel{\"{a}}, M.~Truszczynski, Answer set optimization, in:
  Proceedings of International Joint Conference on Artificial Intelligence
  (IJCAI), 2003, pp. 867--872.

\bibitem{WEGENER2004229}
I.~Wegener, Bdds—design, analysis, complexity, and applications, Discrete
  Applied Mathematics 138~(1) (2004) 229--251.

\bibitem{book-Papadimitriou}
C.~H. Papadimitriou, Computational complexity., Addison-Wesley, 1994.

\bibitem{BrafmanD09}
R.~I. Brafman, C.~Domshlak, Preference handling - an introductory tutorial,
  {AI} Mag. 30~(1) (2009) 58--86.

\bibitem{ramon2020comparison}
Y.~Ramon, D.~Martens, F.~Provost, T.~Evgeniou, A comparison of instance-level
  counterfactual explanation algorithms for behavioral and textual data: Sedc,
  lime-c and shap-c, Advances in Data Analysis and Classification 14 (2020)
  801--819.

\bibitem{ciravegna2020human}
G.~Ciravegna, F.~Giannini, M.~Gori, M.~Maggini, S.~Melacci, Human-driven fol
  explanations of deep learning., in: Proceedings of International Joint
  Conference on Artificial Intelligence (IJCAI), 2020, pp. 2234--2240.

\bibitem{guidotti2018survey}
R.~Guidotti, A.~Monreale, S.~Ruggieri, F.~Turini, F.~Giannotti, D.~Pedreschi, A
  survey of methods for explaining black box models, ACM computing surveys
  (CSUR) 51~(5) (2018) 1--42.

\bibitem{bodria2023benchmarking}
F.~Bodria, F.~Giannotti, R.~Guidotti, F.~Naretto, D.~Pedreschi, S.~Rinzivillo,
  Benchmarking and survey of explanation methods for black box models, Data
  Mining and Knowledge Discovery (2023) 1--60.

\bibitem{WangKB21}
E.~Wang, P.~Khosravi, G.~V. den Broeck, Probabilistic sufficient explanations,
  in: Proceedings of International Joint Conference on Artificial Intelligence
  (IJCAI), 2021, pp. 3082--3088.

\bibitem{ciravegna2023logic}
G.~Ciravegna, P.~Barbiero, F.~Giannini, M.~Gori, P.~Li{\'o}, M.~Maggini,
  S.~Melacci, Logic explained networks, Artificial Intelligence 314 (2023)
  103822.

\bibitem{dervakos2023choose}
E.~Dervakos, K.~Thomas, G.~Filandrianos, G.~Stamou, Choose your data wisely: A
  framework for semantic counterfactuals, arXiv preprint arXiv:2305.17667.

\bibitem{romashov2022baycon}
P.~Romashov, M.~Gjoreski, K.~Sokol, M.~V. Martinez, M.~Langheinrich, Baycon:
  Model-agnostic bayesian counterfactual generator, in: Proceedings of
  International Joint Conference on Artificial Intelligence (IJCAI), 2022, pp.
  23--29.

\bibitem{albini2020relation}
E.~Albini, A.~Rago, P.~Baroni, F.~Toni, Relation-based counterfactual
  explanations for bayesian network classifiers., in: Proceedings of
  International Joint Conference on Artificial Intelligence (IJCAI), 2020, pp.
  451--457.

\bibitem{wu2019counterfactual}
Y.~Wu, L.~Zhang, X.~Wu, Counterfactual fairness: Unidentification, bound and
  algorithm, in: Proceedings of International Joint Conference on Artificial
  Intelligence (IJCAI), 2019, pp. 1438--1444.

\bibitem{DandlCBB23}
S.~Dandl, G.~Casalicchio, B.~Bischl, L.~Bothmann, Interpretable regional
  descriptors: Hyperbox-based local explanations, in: Proceedings of Machine
  Learning and Knowledge Discovery in Databases, Vol. 14171, Springer, 2023,
  pp. 479--495.

\bibitem{kenny2021generating}
E.~M. Kenny, M.~T. Keane, On generating plausible counterfactual and
  semi-factual explanations for deep learning, in: Proceedings of AAAI
  Conference on Artificial Intelligence, 2021, pp. 11575--11585.

\bibitem{arenas2022computing}
M.~Arenas, P.~Barcel{\'o}, M.~Romero~Orth, B.~Subercaseaux, On computing
  probabilistic explanations for decision trees, Proceedings of Advances in
  Neural Information Processing Systems 35 (2022) 28695--28707.

\bibitem{arenas2021foundations}
M.~Arenas, D.~Baez, P.~Barcel{\'o}, J.~P{\'e}rez, B.~Subercaseaux, Foundations
  of symbolic languages for model interpretability, Proceedings of Advances in
  Neural Information Processing Systems 34 (2021) 11690--11701.

\bibitem{EibenOPS23}
E.~Eiben, S.~Ordyniak, G.~Paesani, S.~Szeider, Learning small decision trees
  with large domain, in: Proceedings of International Joint Conference on
  Artificial Intelligence ({IJCAI}), 2023, pp. 3184--3192.

\bibitem{el2023cardinality}
O.~El~Harzli, B.~C. Grau, I.~Horrocks, Cardinality-minimal explanations for
  monotonic neural networks, in: Proceedings of International Joint Conference
  on Artificial Intelligence (IJCAI), 2023, pp. 3677--3685.

\bibitem{MarzariCCF23}
L.~Marzari, D.~Corsi, F.~Cicalese, A.~Farinelli, The {\#}dnn-verification
  problem: Counting unsafe inputs for deep neural networks, in: Proceedings of
  International Joint Conference on Artificial Intelligence ({IJCAI}), 2023,
  pp. 217--224.

\bibitem{ordyniak2023parameterized}
S.~Ordyniak, G.~Paesani, S.~Szeider, The parameterized complexity of finding
  concise local explanations, in: Proceedings of International Joint Conference
  on Artificial Intelligence (IJCAI), 2023, pp. 3312--3320.

\bibitem{BoutilierBDHP04}
C.~Boutilier, R.~I. Brafman, C.~Domshlak, H.~H. Hoos, D.~Poole, Cp-nets: {A}
  tool for representing and reasoning with conditional ceteris paribus
  preference statements, Journal of Artificial Intelligence Research 21 (2004)
  135--191.

\bibitem{RossiVW04}
F.~Rossi, K.~B. Venable, T.~Walsh, mcp nets: Representing and reasoning with
  preferences of multiple agents, in: Proceedings of the Nineteenth National
  Conference on Artificial Intelligence, Sixteenth Conference on Innovative
  Applications of Artificial Intelligence, 2004, pp. 729--734.

\bibitem{AllenGJMR17}
T.~E. Allen, J.~Goldsmith, H.~E. Justice, N.~Mattei, K.~Raines, Uniform random
  generation and dominance testing for cp-nets, Journal of Artificial
  Intelligence Research 59 (2017) 771--813.

\bibitem{GoldsmithLTW08}
J.~Goldsmith, J.~Lang, M.~Truszczynski, N.~Wilson, The computational complexity
  of dominance and consistency in cp-nets, Journal of Artificial Intelligence
  Research 33 (2008) 403--432.

\bibitem{LukasiewiczM19}
T.~Lukasiewicz, E.~Malizia, Complexity results for preference aggregation over
  (\emph{m})cp-nets: Pareto and majority voting, Artificial Intelligence 272
  (2019) 101--142.

\bibitem{buiten2019towards}
M.~C. Buiten, Towards intelligent regulation of artificial intelligence,
  European Journal of Risk Regulation 10~(1) (2019) 41--59.

\bibitem{NYT_Article}
S.~Lohr,
  \href{https://www.nytimes.com/2023/05/25/technology/ai-hiring-law-new-york.html}{A
  hiring law blazes a path for a.i. regulation}, New York Times (May 2023).
\newline\urlprefix\url{https://www.nytimes.com/2023/05/25/technology/ai-hiring-law-new-york.html}

\end{thebibliography}
\clearpage
\appendix

\end{document}